\documentclass[10pt,twocolumn,letterpaper]{article}

\usepackage[pagenumbers]{wacv} % To force page numbers, e.g. for an arXiv version

\usepackage{graphicx}
\usepackage{amsmath}
\usepackage{amssymb}
\usepackage{booktabs}
\usepackage[accsupp]{axessibility} %

\usepackage{lipsum}
\usepackage{booktabs}
\usepackage{algorithm}
\usepackage{algpseudocode}
\usepackage{multirow}
\usepackage{graphicx} 
\usepackage{wrapfig}
\usepackage{enumitem}
\usepackage[dvipsnames]{xcolor}

\usepackage{wrapfig}
\usepackage{makecell}
\usepackage{lscape}

\usepackage[pagebackref,breaklinks,colorlinks]{hyperref}

\usepackage[capitalize]{cleveref}
\crefname{section}{Sec.}{Secs.}
\Crefname{section}{Section}{Sections}
\Crefname{table}{Table}{Tables}
\crefname{table}{Tab.}{Tabs.}

\def\wacvPaperID{2067} % *** Enter the WACV Paper ID here
\def\confName{WACV}
\def\confYear{2025}

\begin{document}

\title{Can Multimodal Large Language Models Truly Perform \\ Multimodal In-Context Learning?}

\author{
Shuo Chen$^{1,3}$ \;
Zhen Han$^{1}$ \; 
Bailan He$^{1,3}$ \; 
Jianzhe Liu $^{4}$ \;
Mark Buckley$^{3}$ \; \\
Yao Qin $^{5}$ \;
Philip Torr$^{2}$ \;
Volker Tresp$^{1,6}$ \;
Jindong Gu$^{2}$\thanks{\vspace{-0.6cm}corresponding author}\;\\
$^{1}$LMU Munich $ $
$^{2}$University of  Oxford $ $
$^{3}$Siemens AG $ $ \\
$^{4}$Technical University of Munich $ $
$^{5}$University of California, Santa Barbara$ $ \\
$^{6}$Munich Center for Machine Learning (MCML)\\
\small\texttt{chenshuo.cs@outlook.com,  jindong.gu@eng.ox.ac.uk}\\}

\maketitle

%%%%%%%%% ABSTRACT
\begin{abstract}
Large Language Models (LLMs) with in-context learning (ICL) ability can quickly adapt to a specific context given a few demonstrations (demos).
Recently, Multimodal Large Language Models (MLLMs) built upon LLMs have also shown multimodal ICL ability, \ie, responding to queries given a few multimodal demos, including images, queries, and answers.
While ICL has been extensively studied on LLMs, its research on MLLMs remains limited. One essential question is whether these MLLMs can truly conduct multimodal ICL, or if only the textual modality is necessary.
We investigate this question by examining two primary factors that influence ICL: 1) Demo content, \ie, understanding the influences of demo content in different modalities. 2) Demo selection strategy, \ie, how to select better multimodal demos for improved performance.
Experiments revealed that multimodal ICL is predominantly driven by the textual content whereas the visual information in the demos has little influence. 
Interestingly, visual content is still necessary and useful for selecting demos to increase performance.
Motivated by our analysis, we propose a simple yet effective approach, termed Mixed Modality In-Context Example Selection (MMICES), which considers both visual and language modalities when selecting demos. 
Extensive experiments are conducted to support our findings and verify the improvement brought by our method. Code is available at \url{https://chenxshuo.github.io/m-icl/}.

\end{abstract}

%%%%%%%%% BODY TEXT

\section{Introduction}
\begin{figure}[t]
  \centering\includegraphics[width=0.92\linewidth]{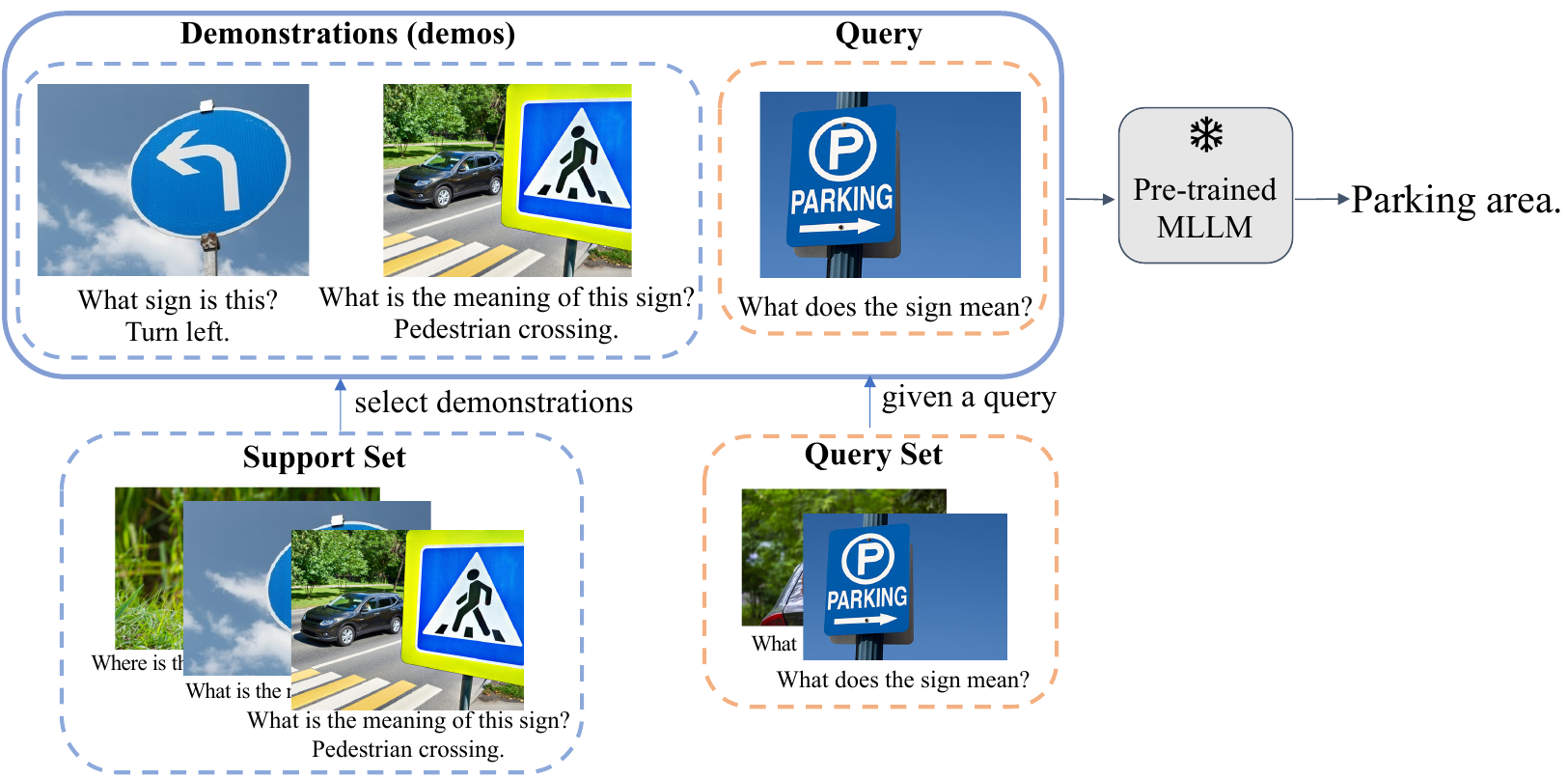}\vspace{-0.1cm}
\caption{In-context learning (2-shot) on visual question answering. Pre-trained MLLMs can perform In-context Learning for a given query based on a few context demos (\ie, a list of images, questions, and answers) selected from a support set.}\vspace{-0.4cm}
\label{fig:icl-example}
\vspace{-0.2cm}
\end{figure}

The in-context learning (ICL) ability of large language models (LLMs) has received great attention~\cite{brown2020language, touvron2023llama, hoffmann2022an, chowdhery2022palm, liang2022holistic}. 
It enables the models to quickly adapt to a specific context given a set of question-and-answer pairs, referred to as demonstrations (demos), without any model parameter updates~\cite{brown2020language, bommasani2021opportunities, dong2022survey}.
Recent Multimodal Large Language Models (MLLMs), which are built upon LLMs, have also displayed multimodal in-context learning (M-ICL) ability~\cite{alayrac2022flamingo, laurenccon2023obelisc, tsimpoukelli2021multimodal, awadalla2023openflamingo, zhao2023mmicl, yu2022generate}. 
These pre-trained models can rapidly adapt to vision-language tasks using few-shot multimodal demos comprising images, queries, and answers. 
For example, as shown in \cref{fig:icl-example}, two images and the corresponding questions and answers are selected as demos from an available support set. Then a pre-trained MLLM generates answers for the query based on the demos. 
Although ICL on LLMs has been intensively explored~\cite{min-etal-2022-rethinking, yoo-etal-2022-ground, wei2023larger, lu-etal-2022-fantastically, liu-etal-2022-makes, an-etal-2023-context}, the understanding of this capability within MLLMs remains largely limited. 
An essential question is \textit{whether these MLLMs can truly perform multimodal ICL, or if only the textual modality is necessary and such in-context learning is still unimodal}. 
This study aims to investigate this question by examining the two main factors influencing the ICL ability, \ie, the demonstration content and the demonstration selection strategy~\cite{xu2024context, dong2022survey}.
% This study focuses on understanding the influence of multimodal content on the ICL ability and the influence of different modalities on the demonstration selection strategy. 
Specifically, we try to answer the following research questions: 
1) How does the demo content in different modalities influence the ICL ability? Does the multimodal ICL rely heavily on the single textual modality?
2) How to select multimodal demos for better ICL performance? Should we rely on images, text, or both when selecting these demos?

For the first question, experiments on multiple MLLMs and vision-language (VL) tasks revealed that textual information is crucial for successful multimodal ICL. In comparison, omitting visual information barely affects the multimodal ICL. 
Specifically, when the images in the demos are removed or replaced with blank images, ICL performance hardly drops. In comparison, language information such as the correct label space and semantics is more crucial than images. Text corruption in the demos can degrade performance significantly. 
To further understand our findings, we analyze the information flow inside the model architecture.
The analysis revealed that the core design of these models, \ie, the masked-cross attention~\cite{alayrac2022flamingo}, can contribute to this phenomenon. 
The masked-cross attention enables the model to receive interleaved image-text sequences but can limit the influence of demonstration images on the ICL performance.
Additional experiments on the model's inner states confirm that the demonstration's visual information hardly impacts the model's output. In contrast, text in the demos has a substantial and direct influence on the model's output.
To conclude, \textit{the content of demonstration images barely influences the ICL ability and the demonstration texts contribute more to the ICL performance}. 

To answer the second question, we investigate demo selection strategies using different modalities. Experiments show that visual information is still necessary and useful to boost multimodal ICL performance. 
Compared to random selection, using image similarity to select demos can improve ICL performance. 
This indicates that although visual content is less significant in ICL, it can influence the demo selection process and improve ICL performance further.
A reason is that visually similar demos are more likely to bring informative textual information, which is helpful for enhanced ICL performance. 
However, we find that selecting solely on visual or language modality can fall short and a better selection strategy should consider both. 
Based on our analysis, we propose a simple yet effective demonstration selection strategy, termed Mixed Modality In-Context Example Selection (MMICES), which considers both visual and language modalities. Concretely, the visual modality is first used to filter potential demo candidates. Then MMICES ranks and selects demos considering the language modality. By factoring in both visual and language information, demos selected by MMICES are related to both the query image and text. 
In conclusion, \textit{although the content of demo images does not influence the ICL ability, the similarity between demo images and the query image can greatly affect the demo selection and thus influence the ICL performance. Therefore, a better demonstration selection strategy should consider both modalities.}

To summarize, our main contributions are as follows:
\begin{itemize}[itemsep=3pt,topsep=2pt,parsep=0pt]
    \item This research investigates whether MLLMs can truly perform multimodal ICL. By examining the influence of multimodal demo content, it is revealed that textual information is more essential than visual information in the demos. Surprisingly, removing images from the demos results in a negligible decline in the ICL performance, whereas corruption of texts leads to a significant performance decrease.
    % \item Our research examines how the multimodal content in the demos influences MLLMs' ICL ability. It is revealed that textual information is more essential than visual information in the demos. Surprisingly, removing images from the demos results in a negligible decline in the ICL performance, whereas corruption of texts leads to a significant decrease.
    \item Our study then explores the demo selection strategy and indicates that a single modality is inadequate for selecting better demonstrations. Both visual and textual modalities should be considered when selection demonstrations to enhance  ICL performance.
    \item Motivated by our analysis, we propose a simple yet effective method, dubbed MMICES, to enhance the ICL performance of pre-trained MLLMs. Extensive experiments show that MMICES outperforms existing demonstration selection methods in various settings. 
\end{itemize}

%Extensive experiments have verified the effectiveness of MMICES on multiple models and various datasets.
%especially, masked cross-attention mechanism~\cite{alayrac2022flamingo}, a widely-adopted approach for incorporating visual data into pre-trained LLMs~\cite{alayrac2022flamingo, awadalla2023openflamingo, laurenccon2023obelisc}. 
%This per-image masking approach can limit the model to only process visual tokens from the most recent image for each text token, rather than all preceding images in the interleaved input sequence. 
%For example, when generating an answer token, the model only directly engages with the visual tokens from the query image.
%The subsequent self-attention mechanism with the demonstration text embeddings indirectly enables its dependence on all prior demonstration image tokens.
%Additional experiments on the model's inner states confirm that the visual data from the demonstrations hardly impacts the model's output. In contrast, text in the demonstrations has a substantial and direct influence on the model's output.

\section{Related Work}
\vspace{0.1mm}
\noindent\textbf{Multimodal In-Context Learning.} Frozen~\cite{tsimpoukelli2021multimodal} is the first attempt for ICL in multimodality by leveraging a frozen GPT-like LM.
Flamingo~\cite{alayrac2022flamingo} demonstrated stronger ICL performance and can handle flexible interleaved text and visual sequences. It utilizes a masked cross-attention mechanism that integrates visual information into pre-trained LLMs and allows any number of visual inputs. This capability makes the ICL possible and many MLLMs are therefore not suitable for ICL such as BLIP~\cite{li2022blip}, MiniGPT~\cite{zhu2023minigpt}, \etc.
OpenFlamingo~\cite{awadalla2023openflamingo} and IDEFICS~\cite{laurenccon2023obelisc} are popular open-source reproductions of Flamingo with competitive ICL performance. 
Otter~\cite{li2023otter} adopts instruction tuning to support more flexible tasks but it is still based on OpenFlamingo. 
Some other works aim to alleviate the dependency on large-scale pre-training~\cite{chen2023sinc, monajatipoor2023metavl, jin2021good, chen2023lightweight}.
However, their performances are not competitive compared to pre-trained MLLMs such as Flamingo.
Some recent models also demonstrate in-context learning ability~\cite{huang2024language, peng2023kosmos, sun2023generative, sun2023generative2, zhao2023mmicl}.
However, Kosmos~\cite{huang2024language, peng2023kosmos} is designed for grounding and referring. Emu~\cite{sun2023generative, sun2023generative2} focuses more on generating outputs varying in modalities, and MMICL~\cite{zhao2023mmicl} shows a limited in-context performance increase compared to the zero-shot setting.
In contrast, we focus on understanding the in-context learning ability of these MLLMs and seek more effective demonstration selection strategies for diverse multimodal tasks.

\vspace{0.2mm}
\noindent\textbf{Understanding In-Context Learning.} 
LLMs have demonstrated impressive ICL ability~\cite{brown2020language, touvron2023llama, hoffmann2022an, chowdhery2022palm}.
A line of research focuses on understanding the influence of demo content on ICL of LLMs~\cite{min-etal-2022-rethinking, yoo-etal-2022-ground, lu-etal-2022-fantastically, liu-etal-2022-makes, an-etal-2023-context}. \cite{min-etal-2022-rethinking} found that the correct input-label mapping is not as important as expected whereas label space exposure and demo distribution have much more influence on the ICL performance. 
\cite{lu-etal-2022-fantastically} demonstrated the influence of order sensitivity on the ICL performance. 
Besides, \cite{an-etal-2023-context} focused on how the demo diversity, similarity, and complexity affect ICL ability. 
Additionally, some works studied how to select better demonstrations to increase ICL performance~\cite{xu2024context}. Various methods have been proposed, such as clustering retrieval~\cite{yu2022generate, zhang2022automatic}, iterative retrieval~\cite{scarlatos2023reticl}, demo retrievers~\cite{wang2023learning, liu2021makes}, \etc.   
However, the understanding of ICL on MLLMs is still underexplored. \cite{yang2023exploring, li2023configure} have explored better in-context configurations but they have not studied the importance of visual and textual content, and only conducted experiments on a single task. 
This study aims to understand both the influence of multimodal demo content and the demo selection strategy on various tasks such as visual question answering, visual reasoning, and image captioning.

\section{Understanding Influence of Demo Content in Multimodal ICL}
\label{sec:understanding}

\vspace{0.1mm}
\noindent\textbf{Multimodal In-Context Learning Formulation on MLLMs.}
An input query $q$ from a query set, \ie, an image $I_q$ and a question/instruction $T_q$, coming after a context prompt $C_q$, is sent to a pre-trained MLLM $f$. The context prompt $C_q$ consists of $N$ task demonstrations from a support set $S$. Each demonstration includes image $I_i$, instruction $T_i$, and response $R_i$. 
Then $f$ generates a response $R_q$ to the input query $q$, \eg, the answer to $T_q$, based on image $I_q$ and the demo context $C_q$. 
Specifically, the ICL can be written as:
% \begin{equation}
%\begin{gathered}
    $R_q=f([C_q,q]),
    \textrm{where} \;\; q=\langle I_q,T_q \rangle, \;\; C_q=\{ \langle I_i, T_i, R_i \rangle \}_N.$
%\end{gathered}
% \end{equation}  

\noindent\textbf{Experimental Setting for M-ICL.} Four popular VL datasets across three VL tasks are applied in this study to evaluate the modality significance, namely VQAv2~\cite{goyal2017making} and OK-VQA~\cite{okvqa} for visual question answering (VQA), GQA~\cite{hudson2019gqa} for visual reasoning, and MSCOCO~\cite{chen2015microsoft} for image captioning. 
We take the Flamingo~\cite{alayrac2022flamingo} as an example and draw similar conclusions from IDEFICS~\cite{laurenccon2023obelisc}. Please refer to the Supplementary Section 3 for more results.

\begin{figure*}[t]
  \centering
  \includegraphics[width=1.0\linewidth]{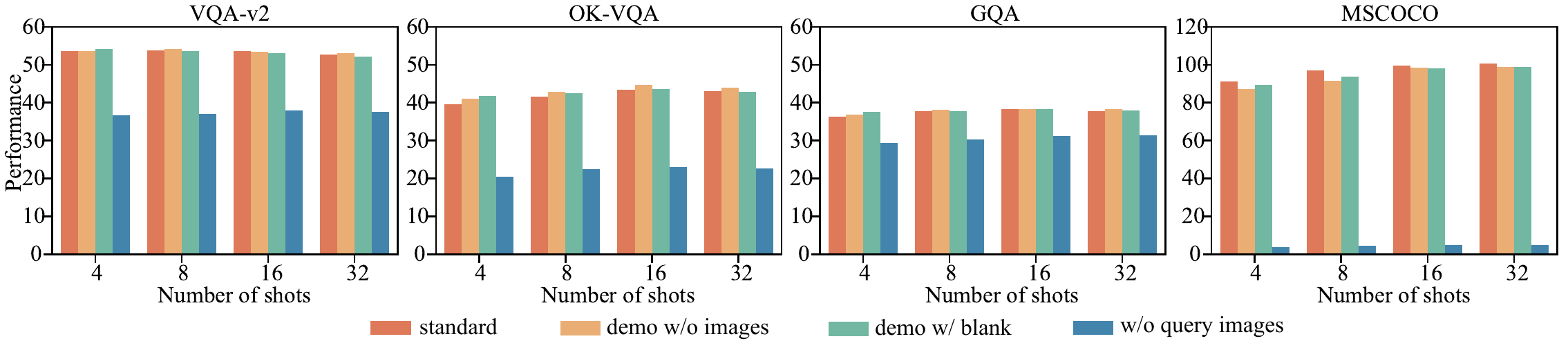}\vspace{-0.4cm}
\caption{The ICL performance is almost the same when removing the visual information in the demonstration.
Compared to the \textit{standard} scenario,  exclusion and replacement of images in the demonstration hardly impact the ICL performance (as shown in the first three bars of each sub-figure). Conversely, the removal of the query image results in substantial performance degradation (as indicated by the last bar in each sub-figure).}
\label{fig:understand-images} 
\vspace{-0.4cm}
\end{figure*}

\subsection{Influence of Visual Information on M-ICL}
\label{sec:understanding-visual}
% \vspace{0.1mm}
% \noindent\textbf{Importance of Visual Information.} 
Unlike ICL on LLMs, ICL in MLLMs incorporates visual information into the demonstration. This visual information can take the form of images used for tasks such as VQA. To evaluate the significance of images in the demonstrations, we have devised the following settings:
\begin{itemize}[itemsep=2pt,topsep=0pt,parsep=0pt]
     \item \textbf{\textit{standard}} setting refers to the scenario where both demonstrations and queries incorporate their respective original image-question pairs.
    \item \textbf{\textit{demo w/o images}} describes the case where all the images in $C$ are removed.
    This results in the context $C$ with $N$ text-only instructions such as the questions in VQA or the captions in the task of image captioning. 
    \item \textbf{\textit{demo w/ blank images}} refers to the scenario where the images and image position tokens in the demos are kept but the original images are replaced with blank ones, \ie, all pixel values are set to $255$. 
    These blank images do not provide any valuable information. 
    \item \textbf{\textit{demo w/o query images}} refers to the setting where the image $I_q$ in the query $Q$ is removed whereas the images in the demonstrations are retained. 

\end{itemize}

\cref{fig:understand-images} presents the ICL performance in different visual demonstration settings, given randomly selected demonstrations.
Compared with the \textit{standard} setting, both the \textit{demo w/o images} and \textit{demo w/ blank} retain most of the ICL performance and some performances remain relatively unchanged. 
Conversely, the \textit{demo w/o query images} setting results in a substantial decline in the ICL performance, with up to a $50\%$ performance drop on VQA and nearly a $100\%$ performance decrease on image captioning. \cref{fig:understand-images} suggests that the visual information in the demonstrations has a minimal impact on the ICL performance. But the images in the query are still important for the inference.

\subsection{Influence of Textual Information on M-ICL}
\label{sec:understanding-textual}
% text information importance 
% \vspace{0.1mm}
% \noindent\textbf{Importance of Textual information.} 
Besides exploring the impact of visual information in the demonstrations, we also assess the significance of textual content using the following settings: %given randomly selected demonstrations 
\begin{itemize}[itemsep=2pt,topsep=0pt,parsep=0pt]
    \item \textbf{\textit{standard}} refers to the case where demos incorporate their respective original image-question pairs. 
    \item \textbf{\textit{different answer for same question}} corresponds to the case where the original answer is replaced with another one from the same question. Despite the question remains the same, the replacement answer can vary due to the differences in the image content.
    \item \textbf{\textit{random question}} describes the case where the original question $T_i$ is replaced with another different $T_j$ but the answer remains unchanged.
    \item \textbf{\textit{random words as labels}} refers to the case where the original $R_i$ in the demo, such as answers in VQA and captions in image captioning, is replaced with random English words. The demo text is hence meaningless. 
\end{itemize}

\begin{figure}[t]
    \centering
    \includegraphics[width=1.0\linewidth]{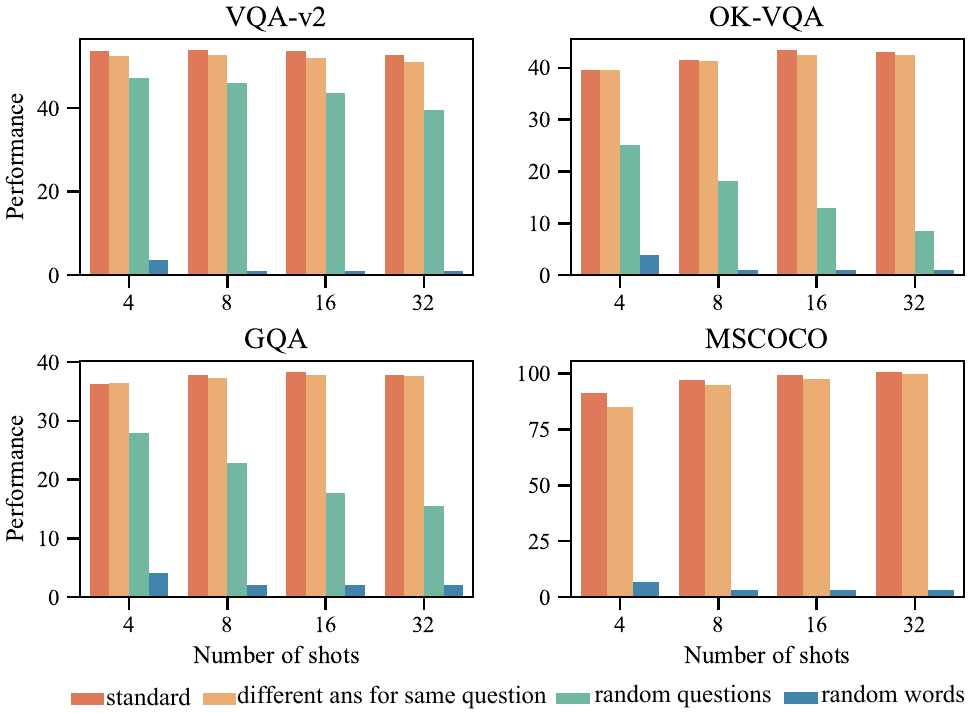}%\vspace{-0.4cm}
    \caption{The ICL performance varies under different text demo settings. Performance in \textit{different answer for same question} can still be maintained (the light orange bars). However, performance significantly decreases in \textit{random question} and \textit{random words as labels} (the green and blue bars).}\vspace{-0.4cm}
\label{fig:understand-language} 
\end{figure}

ICL performance across these settings is displayed in \cref{fig:understand-language}.  Compared to \textit{standard} setting, \textit{random question} leads to a significant drop in performance, and altering labels to random words drastically reduces the performance to nearly zero, as seen in the last bar of each sub-figure. 
When compared to results in \cref{fig:rices-wo-images}, changes in texts can severely affect ICL performance. 
However, \textit{different answer for same question} only marginally impacts performance, regardless of incorrect labels related to the provided query image. This can be because the correct label space and semantics are more crucial than images. Even when answers are inconsistent with images, they can still offer the correct label space and semantics, because they respond to the same question. For example, given an image showing 3 books and the question "How many books are there?", other answers (\eg, 5) to this question still show the correct label space, \ie a number. This finding is consistent with the conclusion from previous experiments, indicating that images have minimal influence on the outcomes. This phenomenon again proves the critical role of texts in the demonstrations.
Hence, we conclude that in MLLMs, the in-context learning ability is primarily driven by textual information, which shows a more substantial influence than visual information in the demonstrations.
\subsection{Further Investigating Content Influence}
\label{sec:attention}

\begin{figure}[t]
  \centering
  \includegraphics[width=0.7\linewidth]{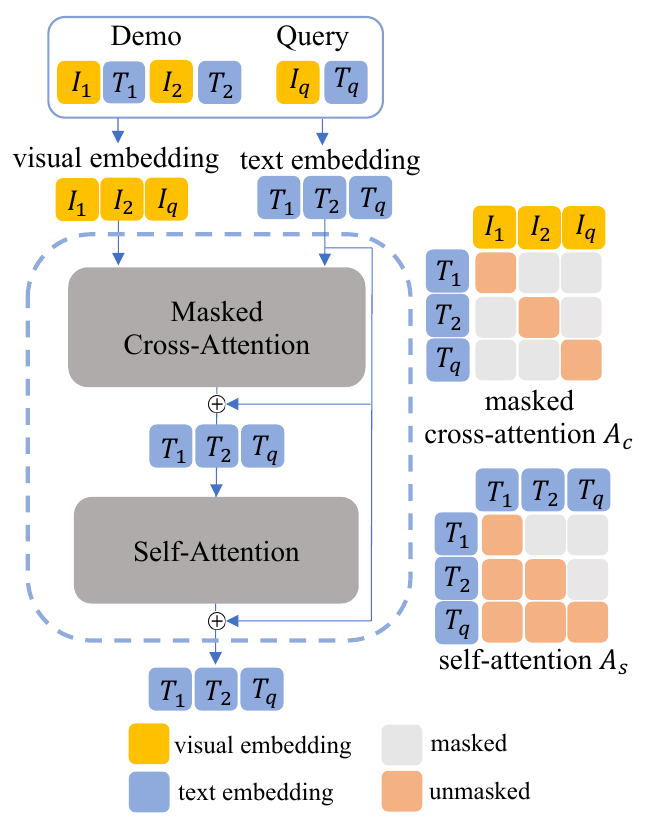} \vspace{-0.2cm}
\caption{Model block supporting interleaved image-text inputs. Visual and language information, \ie, $I$ and $T$, are first fused using a masked cross-attention layer, where each text token is only conditioned on the last preceding image. 
Visual embeddings $\mathbf{I_1}$ and $\mathbf{I_2}$ from demonstration images cannot directly influence query text embedding $\mathbf{T_q}$, and $\mathbf{T_q}$ only sees $\mathbf{I_q}$ in the masked cross-attention, as shown in the last row of $\mathbf{A_c}$.}
\vspace{-0.3cm}
\label{fig:OF-attention}    
\end{figure}

Previous subsections highlight the dominant role of textual information in ICL for MLLMs, yet leave several questions unanswered: 1) Why do the images in demos barely affect the ICL performance? 2) Why is the query image still useful? 3) Why does the textual information dominate the ICL ability? 
Existing literature~\cite{li2023otter, zhao2023mmicl, li2023mimic, doveh2024towards} shows that the pre-training datasets cannot provide sophisticated context information and may limit the ICL ability of the trained MLLMs. 
To further investigate the reasons, this work studies the popular model structures used in these models, \ie, the masked cross-attention architecture~\cite{alayrac2022flamingo}, and analyzes the influence of multimodal demos.

% on the flamingo's mechanism and our guess
The ability to handle interleaved text and image sequences makes ICL possible~\cite{alayrac2022flamingo}.
An illustration is presented in \cref{fig:OF-attention}, with two demos and a query, each of which contains an image and corresponding text such as $I_1$ and $T_1$ in the first demo. 
The masked cross-attention layer enables the language models to incorporate visual information for the next-token prediction. This layer also limits the visual tokens the model can see at each text token. 
Specifically, at a given text token, the model only attends to the visual tokens of the last preceding image, rather than to all previous images in the interleaved sequence. 
For example, text embedding $\mathbf{T_q}$ can only attend to the query image $\mathbf{I_q}$ in the masked cross-attention layer, as shown in the last row of $A_c$ in \cref{fig:OF-attention}.
Therefore, demonstration images $I_1$ and $I_2$ cannot directly pass their visual information to the query text embedding $\mathbf{T_q}$, as $\mathbf{T_q}$ is limited to interacting with the query image representation $\mathbf{I_q}$ in the masked cross-attention layer. 
Only in the subsequent self-attention layer can $\mathbf{T_q}$ indirectly access the information from $I_1$ and $I_2$ through the demo text embeddings $\mathbf{T_1}$ and $\mathbf{T_2}$. Because they have already processed the visual information from $I_1$ and $I_2$ in the masked cross-attention layer. 
We argue that the masked cross-attention mechanism with such per-image attention masking~\cite{alayrac2022flamingo} diminishes text tokens' dependency on all previous images. 
In other words, relying solely on the self-attention layer for transferring visual information to text tokens is difficult.
Thus, it is observed that the generated output tokens primarily focus on the latest image, \ie, the query image, and largely disregard the visual information of the previous images, \ie, the demo images.

% result and visualization 
\begin{figure}[t]
  \centering
  \includegraphics[clip, width=0.95\columnwidth]{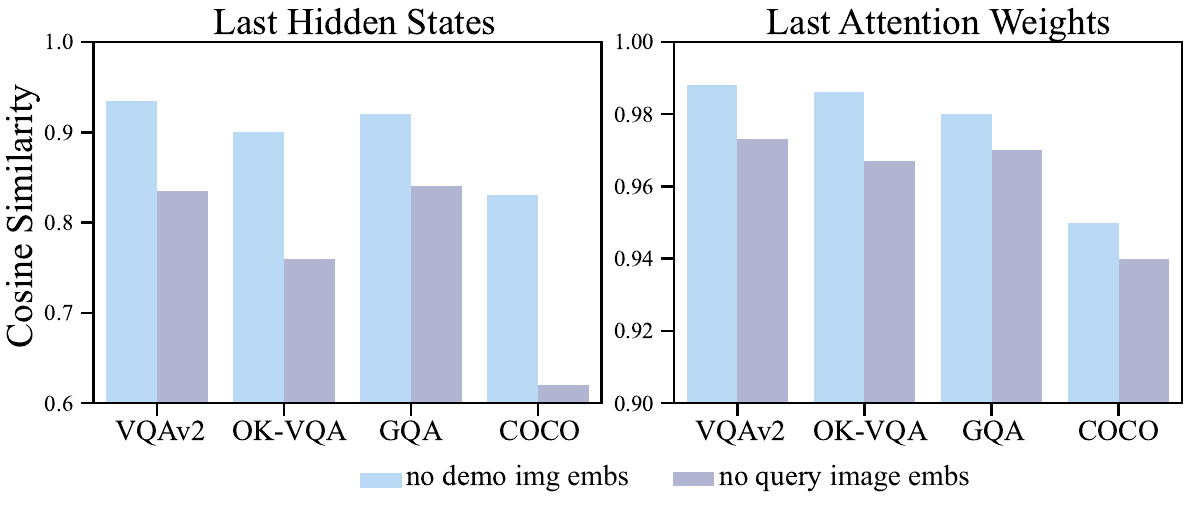}\vspace{-0.2cm}
\caption{The left figure shows the cosine similarity between hidden states in the standard setting and removing images in the demos (blue bars). Grey bars are cosine similarity between standard setting and removing query images. The right figure shows the similarity of the corresponding attention weights in the last decoder layer. Omitting demonstration visual embeddings leads to similar hidden states, but excluding query images increases their dissimilarity, indicating the minimal influence of the demo images.
}\vspace{-0.3cm}
\label{fig:OF-cosine}\vspace{-0.3cm}
\end{figure}

% our method: compare and visualize the attn weights and input, output embeddings w/ and w/o media tokens 
To verify our assumptions, we compare the self-attention weights and self-attention outputs of the language decoder block in the standard setting with two scenarios, \ie, with and without providing visual information in demos. 
If the combination of masked cross-attention across modalities and the self-attention on text tokens maintains the dependency on previous images, excluding visual information from previous demonstration images will lead to different attention behaviors, \eg, different attention weights and hidden states. Otherwise, if the weights and hidden states remain almost the same after removing visual information in the demos, the model does not attend much to previous demonstration images. 
Specifically, we compute the cosine similarity on the last row of hidden states and attention weights in the last decoder layer for each generation forward and then average the results over the whole dataset. To remove the visual information in the demos, we mask the visual embeddings of the demo images, such as $\mathbf{I_1}$ and $\mathbf{I_2}$, by setting the weights to $0$ and keeping the query image embedding $\mathbf{I_q}$. \cref{fig:OF-cosine} presents the results. Removing demonstration visual embeddings leads to around $90\%$ similar hidden states whereas excluding query images makes the hidden states much more dissimilar. 
These differences in similarity confirm our assumption and analysis above, highlighting the insignificance of demo images for existing MLLMs in ICL. 
This popular architecture's inherent limitation could hinder models like Flamingo~\cite{alayrac2022flamingo, awadalla2023openflamingo} and IDEFICS'~\cite{laurenccon2023obelisc} ICL ability. For some newer models that do not use this structure, such as Qwen-VL~\cite{bai2023qwen}, similar patterns are also observed as shown in Supplementary Tab. 9. However, as different factors like training scheme~\cite{doveh2024towards} and dataset design~\cite{zhao2023mmicl} may come into play. Further investigation into these models is part of our future work.

 % and IDEFICS-2~\cite{laurenccon2024matters}
% In summary, this section shows that language information is more significant than visual information in the demos. Excluding images from the demos barely affects the ICL performance whereas corruption of texts and removing query images lead to a significant performance decrease.

% \input{Styles/sections/3-2-new-exps}
% \section{Selecting Better Multimodal demos}
\section{Understanding Demo Selection in M-ICL}
\label{sec:MMICES}

\subsection{Demo Selection using Single Modality}
% callback introduction; 注意介绍方法的时候已经要用加粗 Why visual modality can help，even they have only marginal impact on model output; 要想想 我们的方法怎么根据 上面的analysis motivated 很重要 motivation details 可以再介绍方法的时候再写
% motivation
% logic: images in demos are useless -> similarity retrieval can increase performance because the similar text 
% -> Some demos selected solely on image similarity can be irrelevant. -> also need to consider text information 
\cref{sec:understanding} highlights the influence of text in the demos and this section further investigates how to select effective demos for better multimodal ICL performance. 
Two common approaches to selecting demos are investigated here, \ie, random selection and Retrieval-based In-Context Examples Selection (RICES)~\cite{yang2022empirical, alayrac2022flamingo, awadalla2023openflamingo}. The first randomly selects demos from the support set, disregarding different queries, whereas RICES retrieves demos with similar images by comparing them to the query images.

\begin{figure}[t]
  \includegraphics[clip,width=\columnwidth]{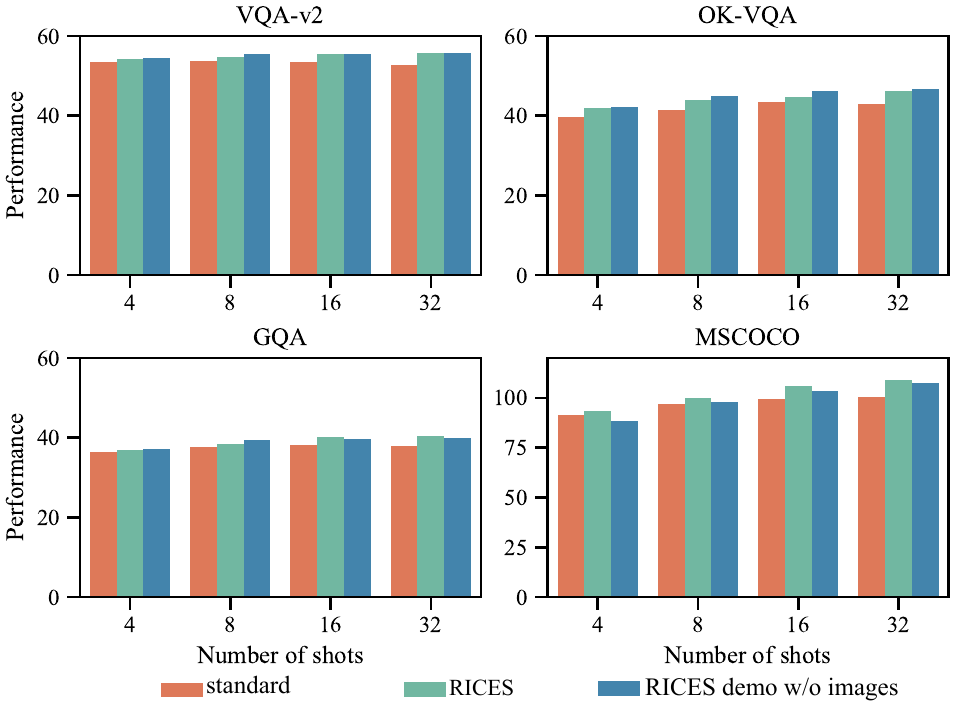}\vspace{-0.4cm}
\caption{The performance of RICES barely changes when removing the images in the demos. RICES leads to better performance (middle bar in each sub-figure) compared to the standard random selection (the first bar). However, disregarding the images in the demos chosen by RICES has a minimal impact on the performance (the last bar) compared to the original RICES.}
\label{fig:rices-wo-images} 
\end{figure}

Compared to random selection, RICES has been proven useful to boost the ICL performance on various tasks~\cite{alayrac2022flamingo, awadalla2023openflamingo}.
If visual information has a minimal impact on ICL performance,  as shown in \cref{sec:understanding}, why does RICES yield better results?  To answer this question, we applied the \textit{demo w/o images} setting to RICES, referred to as \textit{RICES demo w/o images}. It means that the images in the context demos selected by RICES are removed and all the other textual information remains unchanged. The results are presented in \cref{fig:rices-wo-images}. Surprisingly, nearly all of the ICL performances in the \textit{RICES demo w/o images} setting remain relatively unchanged. 
This suggests that images in the selected demos do not significantly contribute to the performance gain. Instead, the remaining textual information plays a more crucial role and this also aligns with our findings in \cref{sec:understanding}.  
Nevertheless, images can still serve as a good criterion for selecting demos. 
This is because the demonstration texts retrieved by RICES contain query-related background information, which is a crucial factor in achieving such performance gain. 
For instance, given a query image depicting a dinner table laden with food, RICES selects demos that are also related to food and dinner. This relevant background knowledge aids the model in better comprehending the context and recalling the necessary information for generating an appropriate response to the query ~\cite{liu-etal-2022-makes, yang2022empirical, dai2023can, olsson2022context}.

\begin{figure}[t]
\includegraphics[clip, width=\columnwidth]{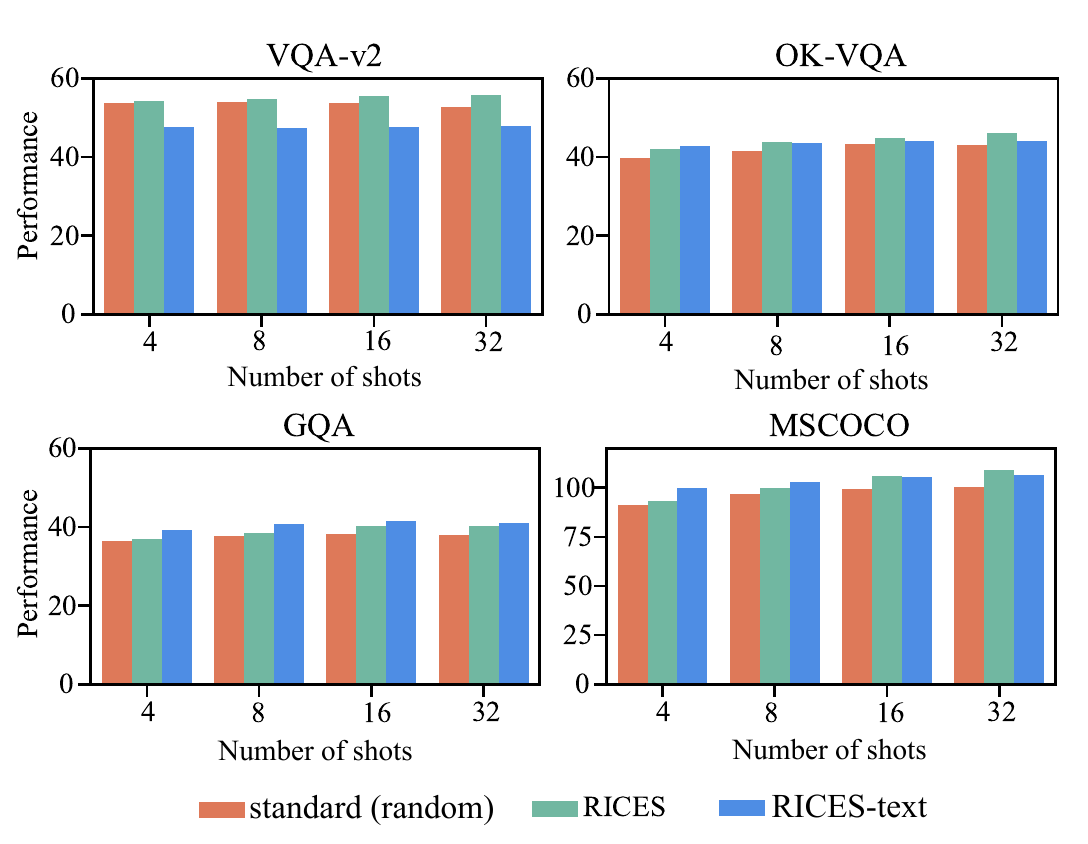}
\vspace{-0.5cm}
\caption{Although textual content in the demos is essential, retrieval based exclusively on text (the last bar in each sub-fig) cannot guarantee a better ICL performance.}
\label{fig:rices-text} 
\end{figure}

However, context demos chosen solely on visual similarity may not always be informative. 
This limitation arises because questions related to visually similar images are not necessarily interconnected. 
In other words, questions in the demos can differ from those in the query despite the images' similarity. 
As demonstrated in the first row of \cref{fig:mmices}, the query question addresses the store selling the pizza, while the retrieved demos ask about the type and shape of the pizza. 
To identify more informative and relevant demos for queries, the retrieval process should not exclusively depend on visual information. 
It should also integrate the available textual information from both demos and queries to find more informative demos.
Despite the importance of textual information, retrieval based exclusively on text presents its own challenges as shown in \cref{fig:rices-text}. The \textit{RICES-text} setting describes the case where demos are selected only using text similarities. The ICL performance of \textit{RICES-text} is lower than RICES on OK-VQA and even worse compared to the random selection on VQAv2. 
This can be attributed to the limitations of general text queries, such as "What is in this picture?". Such queries lack specificity, providing insufficient information and potentially misleading the model generation.
In summary, selecting demos based on a single modality is inadequate to provide informative and suitable demonstrations for the model to perform ICL. %We then explore a better strategy for selecting demos considering more modalities.
\subsection{Mixed Modality in-Context Example Selection}

% our solution 
% \SetKwComment{Comment}{/* }{ */}
% \begin{algorithm}[t]
% \caption{\small Mixed Modality In-Context Example Selection (MMICES)}\label{alg:mm-rices}
% \KwIn{query dataset $Q$, support dataset $S$, vision encoder $E_v$, text encoder $E_t$, $K$, $N$}
% \KwOut{context demos $C$ chosen from $S$ for $Q$}
% Initialization: $C \gets []$\;
% \For{query $q \in Q$}{
%     $v_q$ $\gets$ $E_v(q)$ \; \Comment{This is a comment}
%     $t_q$ $\gets$ $E_t(q)$ \; \Comment{This is a comment}
%     visual similar samples $\gets$ choose $K$ most similar samples from $S$ given $v_q$ \;
%     demos $\gets$ choose $N$ most similar demos from the $K$ samples given $t_q$ \;
%     $C \mathrel{+}=$ demos \;
% }
% \end{algorithm}

\begin{algorithm}[t]
    \caption{\small Mixed Modality In-Context Example Selection (MMICES)}\label{alg:mm-rices}
    \begin{algorithmic}[1]
        \Require query dataset $Q$, support dataset $S$, vision encoder $E_v$, text encoder $E_t$, number of pre-filtered samples $K$, number of demos $N$
        \State $C \gets [\ ]$ \Comment{initialize selected demos}
        \ForAll{query $q \in Q$} \Comment{select demos for each query}
            \State $v_q \gets E_v(q)$ \Comment{obtain visual embedding}
            \State $t_q \gets E_t(q)$ \Comment{obtain text embedding}
            \State $s \gets select(K, S, v_q)$ \Comment{choose $K$ most similar samples from $S$ given $v_q$}
            \State $c \gets select(N, s, t_q)$ \Comment{choose $N$ most similar demos from $s$ given $t_q$}
            \State $C \mathrel{+}= c$ \Comment{add selected demos to $C$}
        \EndFor
        \State \Return context demos $C$ chosen from $S$ for $Q$
    \end{algorithmic}    
\end{algorithm}

\begin{figure}[t]
    \centering
    \includegraphics[width=1.0\linewidth]{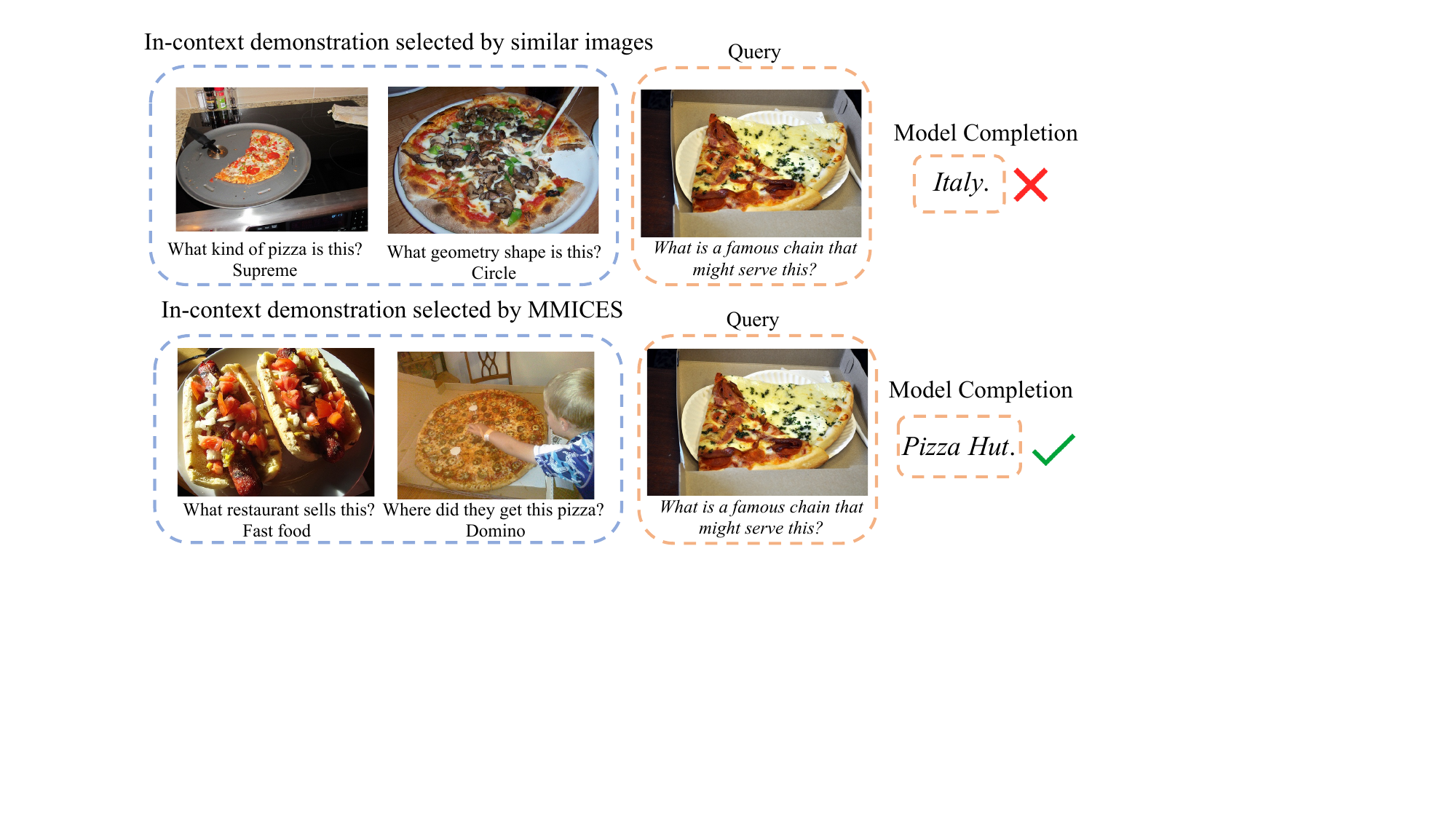}\vspace{-0.4cm}
    \caption{Demos selected by similar images (the top column) and selected by MMICES (the bottom column) given the same query. Demos containing similar images do not necessarily include related textual information to the given query. MMICES considers both visual and language modalities. It provides more informative demos and boosts the ICL performance. More empirical results can be found in Supplementary, Tab. 23.}
    \label{fig:mmices} \vspace{-0.3cm}
\end{figure}

Addressing the above challenges of using a single modality to select demos, we propose to explicitly utilize multimodal information to select demos and design a simple yet effective method, named Mixed Modality In-Context Example Selection (MMICES). 
It initially selects candidates based on image similarity, followed by a reranking based on text similarity, as shown in Alg.~\ref{alg:mm-rices}. 
The objective is to select $N$ context demos for each query $q$ in the query dataset $Q$ (\eg, the test dataset of VQAv2) from the support dataset $S$ (\eg, the training dataset of VQAv2). First, $K$ pre-filtered samples from $S$ are selected based on visual feature similarity. The visual features are extracted from the vision encoder of the MLLM, and $K$ is a hyperparameter.
Then MMICES also considers textual information and selects $N$ most similar ones from the pre-filtered $K$ samples based on textual similarity calculated by a text encoder.
% why does it work; 特别强调 为什么图像没用 我们用图像检索有用 -因为 对应的文字
\textbf{Why is the visual modality in MMICES helpful given its marginal impact in context?}
While the impact of demonstration images has been found to be minimal, they are useful in preliminary demonstration retrieval. \textit{This is mainly because similar images are more likely to bring texts containing related information.}
Therefore, MMICES employs a mixed-modality approach to ensure the provision of high-quality context demos. As illustrated in the second row of \cref{fig:mmices}, the demonstrations selected by MMICES are more closely related to the query and consequently assist the model in generating the correct response.

\section{Experiments}
\label{sec:exp}
\subsection{Experimental Setup}
\label{sec:exp-setting}

\noindent\textbf{Multimodal Large Language Models.}
Seven different models from OpenFlamingo~\cite{awadalla2023openflamingo} (OF) and IDEFICS~\cite{laurenccon2023obelisc} are used in this study. 
The architecture of OF and IDEFICS consists of a frozen large language model with decoder-only structure 
% (\eg, MPT~\cite{mpt7b} in OpenFlamingo and LLaMA~\cite{touvron2023llama} in IDEFICS)
,a frozen visual encoder %(\eg, CLIP-ViT~\cite{radford2021learning}) 
followed by a trainable perceiver resampler~\cite{jaegle2021perceiver, alayrac2022flamingo}. Trainable cross-attention layers are used to bridge the gap between visual and language information. 
Models used in this study vary in their model size (from 3B to 9B), pre-trained datasets, and whether fine-tuned by instruction tuning. 
The instruction-tuned versions are also used in this work, such as  \texttt{IDEFICS-9B-I}.
Besides, experiments are also conducted on some recent models such as MMICL~\cite{zhao2023mmicl}, as described in Supplementary. %, Qwen-vl-chat~\cite{bai2023qwen}, and IDEFICS2~\cite{laurenccon2024matters}. 
% More information is in   %\cref{app:exp-setup}. 

% Both models achieve competitive performance compared to Flamingo~\cite{alayrac2022flamingo}.
%For instance, starts from the base IDEFICS models and is fine-tuned on various datasets, such as M3IT~\cite{li2023m3it} and LLaVA-Instruct~\cite{liu2023llava}. 
% Emu~\cite{yu2022generate} and
% Emu~\cite{yu2022generate}, and MMICL~\cite{zhao2023mmicl}
% The results from OF and IDEFICS are presented in the main paper and please refer to Supplementary Section 4 for more results.
%OpenFlamingo is trained on LAION-2B~\cite{schuhmann2022laion} and Multimodal C4~\cite{zhu2023multimodal}, and IDEFICS is trained on OBELICS~\cite{laurenccon2023obelisc}.

\noindent\textbf{Evaluation Datasets and Metrics.}
Three popular tasks (visual question answering, visual reasoning, and image captioning) and 4 well-known datasets are used. 
For visual question answering, VQAv2~\cite{goyal2017making} and OK-VQA~\cite{okvqa} are adopted. GQA~\cite{hudson2019gqa} is used for visual reasoning and MSCOCO~\cite{chen2015microsoft} for image captioning. Accuracy on the Karpathy-test split is evaluated for VQAv2. For OK-VQA, accuracy on the validation split is evaluated, and accuracy on the test-dev split is used for GQA. CIDEr ~\cite{vedantam2015cider} on the Karpathy-test split is used in MSCOCO.

\subsection{Results}

\begin{table}[t]
\centering
\caption{The performances of random selection (Random), RICES, and MMICES on \texttt{OF-9B}. The highest performance in each scenario is in bold. The results are averaged over 5 evaluation seeds and are reported with their standard deviations. The metric for the MSCOCO is CIDEr. For the other three datasets, top-1 accuracy is reported in percentages.}
\scriptsize
\setlength\tabcolsep{0.1cm}
\begin{tabular}{@{}cccccc@{}}
\toprule
\textbf{Dataset} & \textbf{Method} & \textbf{4-shot} & \textbf{8-shot} & \textbf{16-shot} & \textbf{32-shot} \\ \midrule
\multirow{3}{*}{VQAv2} & Random & 53.52 \scriptsize{$\pm$0.11} & 53.74 \scriptsize{$\pm$0.19} & 53.33 \scriptsize{$\pm$0.26} & 52.38 \scriptsize{$\pm$0.10} \\
 & RICES & \textbf{54.03} \scriptsize{$\pm$0.13} & \textbf{54.67} \scriptsize{$\pm$0.06} & \textbf{55.39} \scriptsize{$\pm$0.12} & \textbf{55.77} \scriptsize{$\pm$0.08} \\
 & MMICES & 53.11 \scriptsize{$\pm$0.03} & 53.56 \scriptsize{$\pm$0.05} & 54.04 \scriptsize{$\pm$0.04} & 55.14 \scriptsize{$\pm$0.02} \\ \midrule
\multirow{3}{*}{OK-VQA} & Random & 39.62 \scriptsize{$\pm$0.29} & 41.56 \scriptsize{$\pm$0.20} & 43.40 \scriptsize{$\pm$0.39} & 42.97 \scriptsize{$\pm$0.11} \\
 & RICES & 42.13 \scriptsize{$\pm$0.13} & 43.87 \scriptsize{$\pm$0.15} & 44.90 \scriptsize{$\pm$0.10} & 46.15 \scriptsize{$\pm$0.06} \\
 & MMICES & \textbf{44.18} \scriptsize{$\pm$0.11} & \textbf{45.61} \scriptsize{$\pm$0.08} & \textbf{46.93} \scriptsize{$\pm$0.08} & \textbf{46.79} \scriptsize{$\pm$0.10} \\ \midrule
\multirow{3}{*}{GQA} & Random & 36.32 \scriptsize{$\pm$0.29} & 37.74 \scriptsize{$\pm$0.32} & 38.28 \scriptsize{$\pm$0.10} & 37.85 \scriptsize{$\pm$0.11} \\
 & RICES & 36.92 \scriptsize{$\pm$0.33} & 38.54 \scriptsize{$\pm$0.14} & 40.16 \scriptsize{$\pm$0.14} & 40.21 \scriptsize{$\pm$0.32} \\
 & MMICES & \textbf{40.73} \scriptsize{$\pm$0.09} & \textbf{41.85} \scriptsize{$\pm$0.10} & \textbf{42.21} \scriptsize{$\pm$0.12} & \textbf{42.07} \scriptsize{$\pm$0.08} \\ \midrule
\multirow{3}{*}{MSCOCO} & Random & 89.82 \scriptsize{$\pm$0.23} & 96.81 \scriptsize{$\pm$0.10} & 99.44 \scriptsize{$\pm$0.19} & 100.53 \scriptsize{$\pm$0.26} \\
 & RICES & 93.45 \scriptsize{$\pm$0.07} & 99.74 \scriptsize{$\pm$0.27} & 105.76 \scriptsize{$\pm$0.03} & 109.12 \scriptsize{$\pm$0.20} \\
 & MMICES & \textbf{100.24} \scriptsize{$\pm$0.20} & \textbf{104.90} \scriptsize{$\pm$0.30} & \textbf{108.66} \scriptsize{$\pm$0.17} & \textbf{109.64} \scriptsize{$\pm$0.24} \\ \bottomrule
\end{tabular}
\label{table:mm-rices-of}
\end{table}

\begin{table}[t]
\centering
\caption{The performances of random selection (Random), RICES, and MMICES on \texttt{IDEFICS-9B}. MMICES achieves the best ICL performance in all settings.}
% \resizebox{0.7\columnwidth}{!}{

\scriptsize
\setlength\tabcolsep{0.1cm}
\begin{tabular}{@{}cccccc@{}}
\toprule
\textbf{Dataset} & \textbf{Method} & \textbf{4-shot} & \textbf{8-shot} & \textbf{16-shot} & \textbf{32-shot} \\ \midrule
\multirow{3}{*}{VQAv2} & Random & 54.90 \scriptsize{$\pm$0.05} & 56.16 \scriptsize{$\pm$0.02} & 56.93 \scriptsize{$\pm$0.18} & 57.21 \scriptsize{$\pm$0.17} \\
 & RICES & 54.79 \scriptsize{$\pm$0.09} & 56.45 \scriptsize{$\pm$0.05} & 57.49 \scriptsize{$\pm$0.06} & 58.52 \scriptsize{$\pm$0.02} \\
 & MMICES & \textbf{56.15} \scriptsize{$\pm$0.01} & \textbf{58.17} \scriptsize{$\pm$0.03} & \textbf{59.23} \scriptsize{$\pm$0.01} & \textbf{59.69} \scriptsize{$\pm$0.02} \\ \midrule
\multirow{3}{*}{OK-VQA} & Random & 49.24 \scriptsize{$\pm$0.22} & 49.54 \scriptsize{$\pm$0.12} & 50.89 \scriptsize{$\pm$0.12} & 51.86 \scriptsize{$\pm$0.12} \\
 & RICES & 48.82 \scriptsize{$\pm$0.02} & 50.55 \scriptsize{$\pm$0.05} & 52.42 \scriptsize{$\pm$0.03} & 53.22 \scriptsize{$\pm$0.04} \\
 & MMICES & \textbf{49.63} \scriptsize{$\pm$0.02} & \textbf{52.16} \scriptsize{$\pm$0.03} & \textbf{53.65} \scriptsize{$\pm$0.07} & \textbf{54.16} \scriptsize{$\pm$0.05} \\ \midrule
\multirow{3}{*}{GQA} & Random & 39.35 \scriptsize{$\pm$0.26} & 40.54 \scriptsize{$\pm$0.17} & 41.38 \scriptsize{$\pm$0.18} & 41.86 \scriptsize{$\pm$0.13} \\
 & RICES & 39.86 \scriptsize{$\pm$0.13} & 41.27 \scriptsize{$\pm$0.29} & 42.65 \scriptsize{$\pm$0.21} & 43.67 \scriptsize{$\pm$0.19} \\
 & MMICES & \textbf{42.66} \scriptsize{$\pm$0.05} & \textbf{44.22} \scriptsize{$\pm$0.08} & \textbf{45.19} \scriptsize{$\pm$0.05} & \textbf{45.36} \scriptsize{$\pm$0.09} \\ \midrule
\multirow{3}{*}{MSCOCO} & Random & 96.45 \scriptsize{$\pm$0.36} & 100.85 \scriptsize{$\pm$0.36} & 103.96 \scriptsize{$\pm$0.38} & 105.02 \scriptsize{$\pm$0.43} \\
 & RICES & 91.20 \scriptsize{$\pm$0.10} & 102.58 \scriptsize{$\pm$0.15} & 108.93 \scriptsize{$\pm$0.10} & 111.02 \scriptsize{$\pm$0.08} \\
 & MMICES & \textbf{101.13}\scriptsize{$\pm$0.12} & \textbf{109.31 }\scriptsize{$\pm$0.09} & \textbf{112.72 }\scriptsize{$\pm$0.05} & \textbf{113.37} \scriptsize{$\pm$0.09} \\ \bottomrule
\end{tabular}

\label{table:mm-rices-idefics}
\end{table}

\begin{table}[t]
\centering
\caption{The performances of random selection, RICES, and MMICES on VQAv2 on OpenFlamingo and IDEFICS. MMICES achieves the best performance in most cases}
% \resizebox{0.7\columnwidth}{!}{
\scriptsize
\setlength\tabcolsep{0.1cm}
\begin{tabular}{@{}cccccc@{}}
\toprule
\textbf{Model} & \textbf{Method} & \textbf{4-shot} & \textbf{8-shot} & \textbf{16-shot} & \textbf{32-shot} \\ \midrule
\multirow{3}{*}{\texttt{OF-3B}} & Random & 44.79\scriptsize{$\pm$0.12} & 45.05\scriptsize{$\pm$0.05} & 45.30\scriptsize{$\pm$0.17} & 45.64\scriptsize{$\pm$0.20} \\
 & RICES & 44.64\scriptsize{$\pm$0.09} & 45.71\scriptsize{$\pm$0.12} & 46.30\scriptsize{$\pm$0.03} & 47.48\scriptsize{$\pm$0.05} \\
 & MMICES & \textbf{47.00}\scriptsize{$\pm$0.06} & \textbf{48.46}\scriptsize{$\pm$0.07} & \textbf{49.50}\scriptsize{$\pm$0.06} & \textbf{49.68}\scriptsize{$\pm$0.03} \\ \midrule
% \multirow{3}{*}{OF-3BI} & Random & 45.54\scriptsize{$\pm$0.12} & 45.77\scriptsize{$\pm$0.19} & 45.71\scriptsize{$\pm$0.15} & 45.05\scriptsize{$\pm$0.19} \\
%  & RICES & 45.06\scriptsize{$\pm$0.09} & 45.41\scriptsize{$\pm$0.07} & 45.65\scriptsize{$\pm$0.04} & 46.11\scriptsize{$\pm$0.12} \\
%  & MMICES & \textbf{48.41}\scriptsize{$\pm$0.01} & \textbf{48.38}\scriptsize{$\pm$0.05} & \textbf{48.96}\scriptsize{$\pm$0.05} & \textbf{48.86}\scriptsize{$\pm$0.04} \\ \midrule
\multirow{3}{*}{\texttt{OF-4B}} & Random & 47.74\scriptsize{$\pm$0.24} & 47.10\scriptsize{$\pm$0.04} & 44.32\scriptsize{$\pm$0.12} & 41.88\scriptsize{$\pm$0.25} \\
 & RICES & 47.70\scriptsize{$\pm$0.04} & 46.68\scriptsize{$\pm$0.18} & 44.91\scriptsize{$\pm$0.07} & 42.86\scriptsize{$\pm$0.08} \\
 & MMICES & \textbf{48.89}\scriptsize{$\pm$0.04} & \textbf{48.61}\scriptsize{$\pm$0.09} & \textbf{46.45}\scriptsize{$\pm$0.07} & \textbf{43.73}\scriptsize{$\pm$0.06} \\ \midrule
% \multirow{3}{*}{OF-4BI} & Random & 47.74\scriptsize{$\pm$0.11} & 46.20\scriptsize{$\pm$0.15} & 44.01\scriptsize{$\pm$0.23} & 46.33\scriptsize{$\pm$0.14} \\
%  & RICES & 48.24\scriptsize{$\pm$0.08} & 46.27\scriptsize{$\pm$0.12} & 44.32\scriptsize{$\pm$0.13} & 47.55\scriptsize{$\pm$0.12} \\
%  & MMICES & \textbf{49.03}\scriptsize{$\pm$0.04} & \textbf{48.22}\scriptsize{$\pm$0.07} & \textbf{47.42}\scriptsize{$\pm$0.03} & \textbf{48.85}\scriptsize{$\pm$0.05} \\ \midrule
\multirow{3}{*}{\texttt{OF-9B}} & Random & 53.52\scriptsize{$\pm$0.11} & 53.74\scriptsize{$\pm$0.19} & 53.33\scriptsize{$\pm$0.26} & 52.38\scriptsize{$\pm$0.10} \\
 & RICES & \textbf{54.03}\scriptsize{$\pm$0.13} & \textbf{54.67}\scriptsize{$\pm$0.06} & \textbf{55.39}\scriptsize{$\pm$0.12} & \textbf{55.77}\scriptsize{$\pm$0.08} \\
 & MMICES & 53.11\scriptsize{$\pm$0.03} & 53.56\scriptsize{$\pm$0.05} & 54.04\scriptsize{$\pm$0.04} & 55.14\scriptsize{$\pm$0.02} \\ \midrule
\multirow{3}{*}{\texttt{IDEFICS-9B}} & Random & 54.90\scriptsize{$\pm$0.05} & 56.16\scriptsize{$\pm$0.02} & 56.93\scriptsize{$\pm$0.18} & 57.21\scriptsize{$\pm$0.17} \\
 & RICES & 54.79\scriptsize{$\pm$0.09} & 56.45\scriptsize{$\pm$0.05} & 57.49\scriptsize{$\pm$0.06} & 58.52\scriptsize{$\pm$0.02} \\
 & MMICES & \textbf{56.15}\scriptsize{$\pm$0.01} & \textbf{58.17}\scriptsize{$\pm$0.03} & \textbf{59.23}\scriptsize{$\pm$0.01} & \textbf{59.69}\scriptsize{$\pm$0.02} \\ \bottomrule
% \multirow{3}{*}{IDEFICS-9BI} & Random & 63.94\scriptsize{$\pm$0.13} & 64.43\scriptsize{$\pm$0.14} & 64.64\scriptsize{$\pm$0.10} & 64.87\scriptsize{$\pm$0.09} \\
%  & RICES & \textbf{64.13}\scriptsize{$\pm$0.08} & \textbf{64.69}\scriptsize{$\pm$0.03} & 65.11\scriptsize{$\pm$0.05} & 65.22\scriptsize{$\pm$0.03} \\
%  & MMICES & 63.51\scriptsize{$\pm$0.13} & 64.46\scriptsize{$\pm$0.04} & \textbf{65.26}\scriptsize{$\pm$0.04} & \textbf{65.5}\scriptsize{$\pm$0.02} \\ \bottomrule
\end{tabular}
\label{table:mm-rices-vqa}
\end{table}

% The MMICES performance is presented . 
MMICES outperforms random selection and RICE across almost all datasets on both OpenFlamingo and IDEFICS, as shown in \cref{table:mm-rices-of} and \cref{table:mm-rices-idefics}. MMICES consistently boosts the ICL performance on OpenFlamingo across various tasks. 
On GQA, MMICES with only $4$ shots ($40.73\%$) is better than the $32$-shot random selection ($37.85\%$) and $32$-shot RICES ($40.35\%$). 
MMICES is also consistently better on OK-VQA where given only $8$ context examples, the performance (\ie, $45.5\%$) is better than random $32$ shots ($42.97\%$) and RICES's $16$ shots ($44.70\%$).  The performance gain is also evident on \texttt{IDEFICS-9B} across all datasets.
For instance, MMICES increases the accuracy on GQA by around $10\%$ given $8$ context examples (from $40.54\%$ to $44.22\%$) compared to random selection and by around $7\%$ compared to RICES (from $41.27\%$ to $44.22\%$). Besides, MMICES achieves comparable accuracy on GQA given only $4$ demonstrations compared to RICES with $16$ demonstrations. 
All the $16$-shot performances from MMICES are higher compared to $32$-shot random selection and $32$-shot RICES, which indicates that with only half of the context examples, MMICES achieves even better results. 
On VQAv2, although the performances on \texttt{OF-9B} do not outperform the RICES, MMICES on all other models still achieves better results (\cref{table:mm-rices-vqa}). 
Overall, MMICES achieves consistently better performance compared to random selection and RICES across models and datasets. 
% demonstrating its effectiveness.

% The reasons can be attributed to that the \texttt{OF-9B} gets affected by the shortcut patterns in the similar demonstrations selected by MMICES\TODO{more analysis?}. 

\begin{figure}[t]
  \centering
  \includegraphics[width=1.0\linewidth]{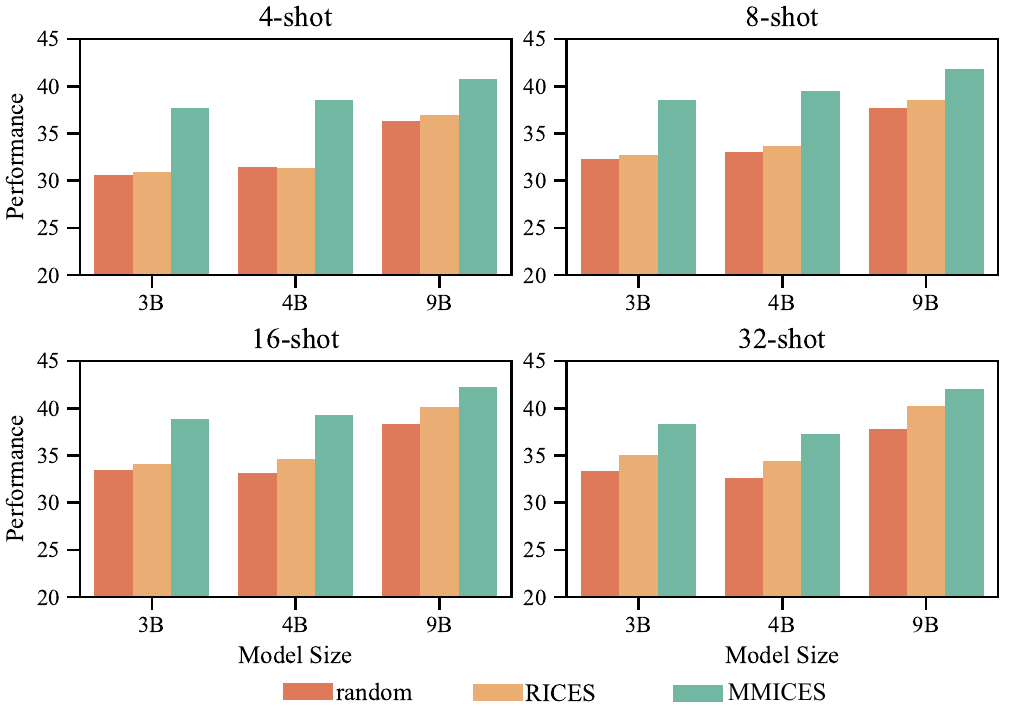}\vspace{-0.4cm}
\caption{MMICES consistently enhances the ICL performance across models of varying sizes. MMICES on smaller models can even outperform RICES on larger models. Results here are from GQA and more results are in Supplementary Section 4.}\vspace{-0.2cm}
\label{fig:ablation-size}
\end{figure}

\begin{figure}[t]
  \centering
  \includegraphics[width=1.0\linewidth]{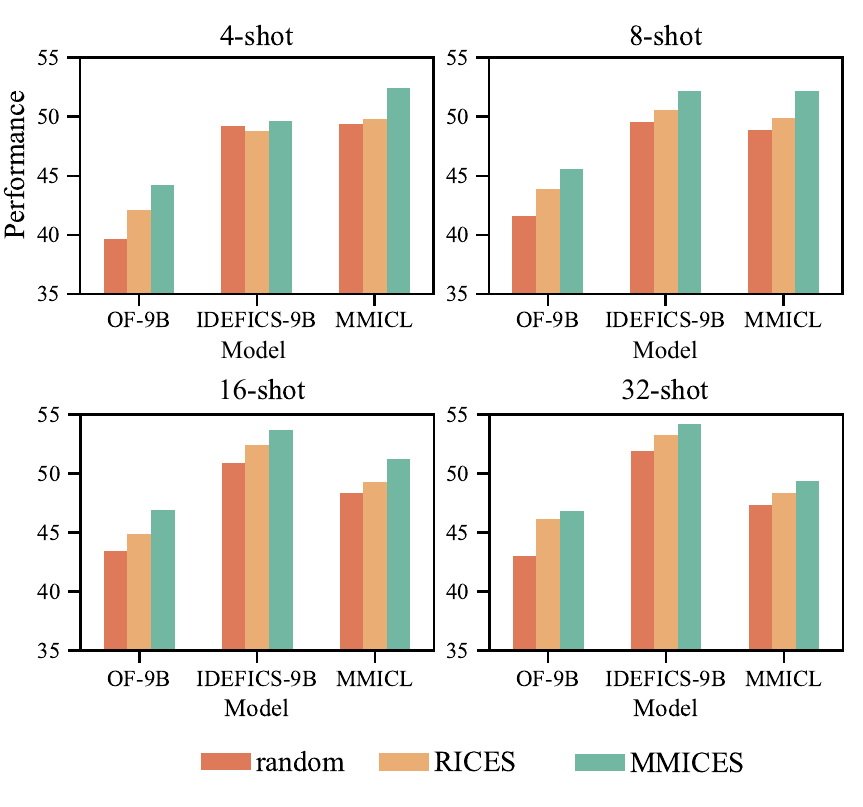} \vspace{-0.4cm}
\caption{The performance of ICL (on OK-VQA) is consistently enhanced by MMICES across different models, including OpenFlamingo~\cite{awadalla2023openflamingo}, IDEFICS~\cite{laurenccon2023obelisc}, and MMICL~\cite{zhao2023mmicl}.}
\label{fig:ablation-model} \vspace{-0.4cm}
\end{figure}

% \subsection{Impact of Underlying MLLMs}
%% Discussion 
We have also conducted extensive experiments on different MLLMs with varying sizes. \cref{fig:ablation-size} presents a performance comparison on the GQA dataset across models from \texttt{OF-3B} to \texttt{OF-9B}. MMICES consistently outperforms RICES by a notable margin. 
It is worth mentioning that MMICES on smaller-size models can achieve better performance, compared to larger-size models using RICES and random selection, especially in $4$ and $8$-shot settings. 
Moreover, the performance gained is consistent across different MLLMs as shown in \cref{fig:ablation-model}. 
% More results and analysis are in Supplementary Section 5. % \cref{app:analysis}

% Regarding the Investigation 
% Sec-3 related exps are also available for IDEFICES and can be shown here if needed

\vspace{-0.1cm}
\section{Conclusion}
This study explores the multimodal in-context learning capability of multimodal large language models and asks whether these models can truly perform multimodal in-context learning.
We first investigate the influence of multimodal demo content on ICL performance and find that \textit{the visual information in the demonstrations has a minimal impact on the ICL performance, while the text is much more important for in-context learning}. 
This work further explores the demo selection strategy and shows that \textit{visual information is still useful for demonstration selection. Besides, both visual and language modalities are necessary to select demonstrations}. 
Based on our analysis, we propose selecting demos based on both visual and text modalities and have designed the Mixed Modality In-Context Example Selection (MMICES) algorithm. Despite its simplicity, MMICES outperforms existing in-context example selection methods across various models and datasets.
% We believe this study can help the community better understand the in-context learning ability of MLLMs.

\vspace{-0.1cm}
\section*{Acknowledgment}
This paper is supported by the DAAD program Konrad Zuse Schools of Excellence in Artificial Intelligence, sponsored by the Federal Ministry of Education and Research.
This work is partially supported by the UKRI grant: Turing AI Fellowship EP/W002981/1 and
EPSRC/MURI grant: EP/N019474/1. We would also like to thank the Royal Academy of Engineering. 
%We also gratefully acknowledge the computational and data resources provided by the
%Leibniz Supercomputing Centre (www.lrz.de) and MCML.

\clearpage
\newpage

%%%%%%%%% REFERENCES
{\small
\bibliographystyle{ieee_fullname}
\bibliography{egbib}
}

\end{document}